# Sub-Setting Algorithm for Training Data Selection in Pattern Recognition

*Gaurav Arwade*, and Sigurdur Olafsson, *Department of Industrial and Manufacturing Systems Engineering, Iowa State University, Ames, USA*

***Abstract*— Modern pattern recognition tasks use complex algorithms that take advantage of large datasets to make more accurate predictions than traditional algorithms such as decision trees or k-nearest-neighbor better suited to describe simple structures. While increased accuracy is often crucial, less complexity also has value. This paper proposes a training data selection algorithm that identifies multiple subsets with simple structures. A learning algorithm trained on such a subset can classify an instance belonging to the subset with better accuracy than the traditional learning algorithms. In other words, while existing pattern recognition algorithms attempt to learn a global mapping function to represent the entire dataset, we argue that an ensemble of simple local patterns may better describe the data. Hence the sub-setting algorithm identifies multiple subsets with simple local patterns by identifying similar instances in the neighborhood of an instance. This motivation has similarities to that of gradient boosted trees but focuses on the explainability of the model that is missing for boosted trees. The proposed algorithm thus balances accuracy and explainable machine learning by identifying a limited number of subsets with simple structures. We applied the proposed algorithm to the international stroke dataset to predict the probability of survival. Our bottom-up sub-setting algorithm performed on an average 15% better than the top-down decision tree learned on the entire dataset. The different decision trees learned on the identified subsets use some of the previously unused features by the whole dataset decision tree, and each subset represents a distinct population of data.**

Gaurav Arwade is a Masters's candidate at, Department of Industrial and Manufacturing Systems Engineering, Iowa State University, Ames, IA, 50011. USA (e-mail: gbarwade@iastate.edu).

Dr. Sigurdur Olafsson is an Associate Professor at, Department of Industrial and Manufacturing Systems Engineering, Iowa State University, Ames, IA, 50011. USA (e-mail: olafsson@iastate.edu).

## I. Introduction

DATA mining is semiautomated process of extracting meaningful and previously unknown patterns from large datasets [1]. Generally, the mined information is stored in the form of a learned model and is used for inference. In recent years a large amount of data has been available easily. Many complex algorithms, such as random forest, support vector machines, and neural networks, are used to achieve good performance in these complex datasets, and typically such methods benefit from more training data. However, sometimes not all the data points are required to make a successful prediction, and it is of inherent interest to understand which data points are critical to learning to predict. In learning theory, the fundamental question is how to characterize algorithmically form a set of critical samples that best describes an unknown model [2].

In classification, a function $\boldsymbol{f}$ given by $\boldsymbol{y} = \boldsymbol{f}(\boldsymbol{x}) + \boldsymbol{\epsilon}$ classifies the $\boldsymbol{y}$ target variable given the set of $\boldsymbol{x}$ predictor variables, and $\boldsymbol{\epsilon}$ is an irreducible random error [3]. The learned function achieves practically the classification task $\hat{\boldsymbol{y}} = \hat{\boldsymbol{f}}(\boldsymbol{x})$ where $\hat{\boldsymbol{f}}$ is an estimate of $\boldsymbol{f}$ and $\hat{\boldsymbol{y}}$ is estimated prediction $\boldsymbol{y}$. The accuracy of $\hat{\boldsymbol{y}}$ depends upon reducible and non-reducible error; accurate estimation of $\hat{\boldsymbol{f}}$ reduces the reducible error, but the irreducible error is an inherent part of the system. Suppose a complex non-linear relationship is present between the target and predictor variables; in such a case, the mapped function may not be sufficiently accurate to achieve good classification accuracy, which is the case most of the time. All the data mining algorithms attempt to learn an underlying universal data distribution function. The question is if the data is best described as all the data points belonging to the same distribution? Practically training datasets are complex as different data points may be drawn from different distributions, so data mining algorithms must learn a complex universal mapping function that may not generalize well. As data space is a mixture of data points from different data distributions, the amount of data space expressed by the model is dependent on the model complexity; a simple model will address a small portion of the data space as it assumes a specific structure of the data whereas complex models express a wide range of data patterns [4]. As shown in fig 1.1a, the complex model $\boldsymbol{Mc}$ tries to learn a universal function that will address the whole data space, but as we can see, it does not cover and generalize over the entire data space well. On the other hand, in fig 1.1b, if we learn multiple simple and unique structure models ***Msi***, we can cover the whole data space with better generality, coverage, and accuracy.

Suppose one identifies different subsets with data points like one another in such cases, the data mining model may learn more than one but simple functions. An ensemble of such functions might explain the data better than a single complex universal function. Thus, besides selecting appropriate data mining algorithms and their parameters, the training data selection also plays a crucial role in the model's performance. Instance selection or instance-based learning is one of the most popular training data selection methodologies.

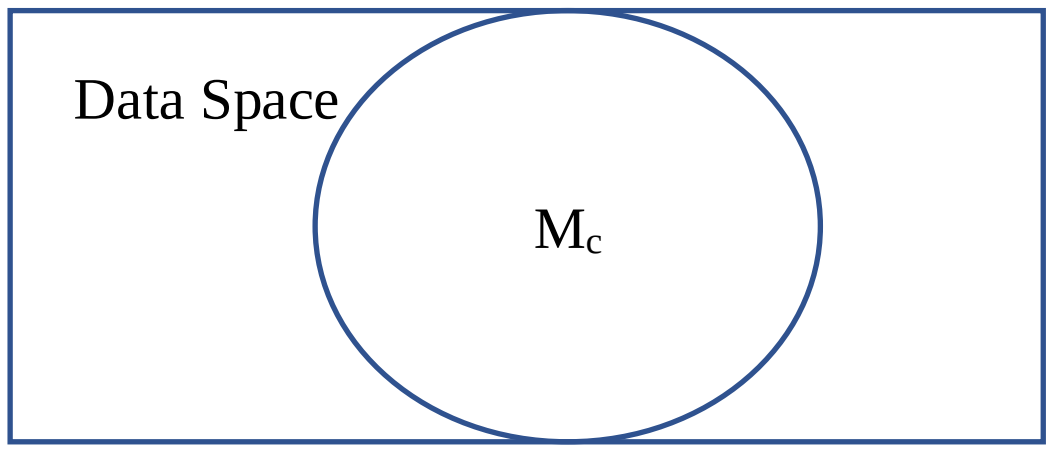

Fig 1.1a [4]

Fig. 1.1a – Data space expressed by a complex model.

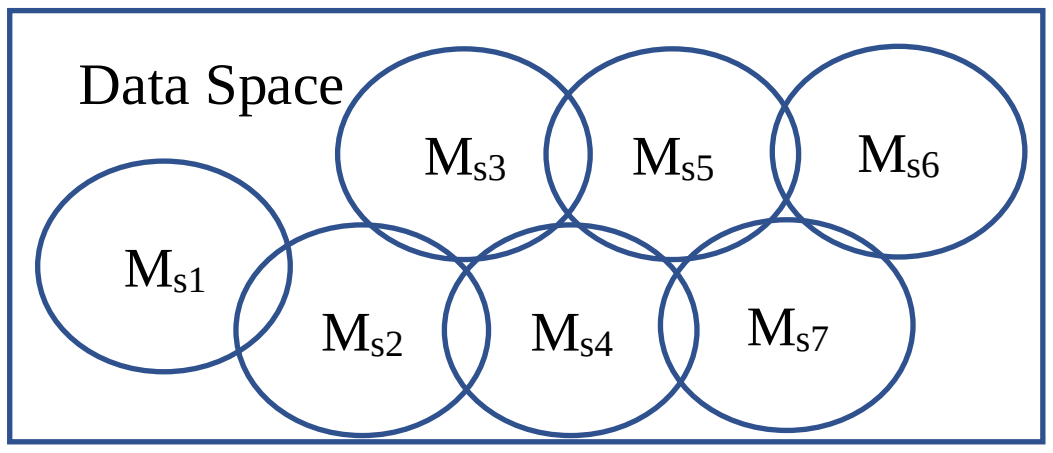

Fig 1.1b

Fig. 1.1b- Data space expressed by multiple simple models.

The main goal of instance-based learning is to find a subset of the dataset to achieve the main aim of the data mining algorithm as if the whole dataset is used. The instance-based learning algorithms achieve this goal by removing redundant, noisy, or superfluous instances from the dataset, improving prediction accuracy, and reducing training time [5]. Some of the earliest work in the field of instance selection was condensed nearest neighbor (CNN) [6], reduced the nearest neighbor (RNN) [7], edited nearest neighbor (ENN) [8], and selective nearest neighbor (SNN) [9]. These early instance selection methods use different nearest neighbor rules to remove noise and data points that don't contribute to learning the mapping functions. More recent training data selection evaluates the density of the instances to select the useful instances. Joel and Mara (2016) propose a CDIS algorithm that evaluates instances of each class separately and keeps only the densest instances in the neighborhood [10]. George and Rodolfo (2020) develop a ranking-based instance selection method RIS1, RIS2 and, RIS3. This algorithm ranks the instances present in the region of a homogenous class higher and eliminates border instances. Some of the latest works tackle training data selection problems using optimization and local search algorithms. Ghadikolaei et al. (2019) proposed a mixed-integer non-linear program that alternates between data selection and function approximation [2]. Lin et al. (2021) use a hybrid memetic algorithm with variable local search [11], whereas Neri and Triguero (2020) use surrogate local search for instance reduction [12].

All the above training data selection methods use different ways to select the data, but all the algorithms aim to compress the training data by removing noise and redundant data points without deteriorating the performance. This paper looks at the instance selection task from a different perspective. We propose a training data selection algorithm identifying multiple subsets of the training datasets with simple patterns, and an ensemble of such patterns can represent the data better. Instead of deleting the instances unhelpful for learning global mapping function, our algorithm learns simple local approximations with previously unidentified structures. This improves the performance of the pattern recognition algorithm as identified simple patterns are more accurate than a global approximation.

In existing literature boosting is a methodology that attempts to exploit different structures in the dataset. Instead of identifying multiple subsets with simple structures boosting achieves good classification performance by learning series of weak learners who are experts in a part of the data and combining them to form a strong learner. A weak learner is a simple mapping function that does better than random guessing [13]. Boosting achieves good performance by increasing the variance of the model that is fitting closer to data. The underlying idea of boosting is to combine simple "rules" to form an ensemble such that the performance of the single ensemble member is improved [14].

Yorav and Robert (1996) proposed the Adaboost algorithm, which performs the weighted addition of the learned weak learners to predict the output. The Adaboost uses a set of weights for training instances, and the weights of the misclassified instances in previous iterations are increased while weights of the correctly classified instances are reduced [15]. Friedman (1999) came up with a gradient boosting machine, an ensemble of the trees, and weighted aggregation of the leaves is performed to predict the output. Unlike Adaboost, in gradient boosting, the weights of the tree leaves and tree structure of the next iteration are learned greedily using the gradient of the loss of the previously learned trees [16]. Chen and Guestrin (2016) proposed XGBoost - a scalable tree boosting system to address the scalability of the gradient boosting machine and added additional regularization to address the overfitting [17].

This paper aims to identify subsets with a unique and simple structure. Although boosting algorithms learn the number of weak learners, they don't align with our idea of simple structures. Even though discovered structures in boosting have learning bias towards certain instances, they are global as they are learned using the entire dataset. The learned weak learners have a high weight in a part of the data and low weight in the remaining data, but this weight is never zero. All the week learners contribute to the prediction of every instance, no matter how small the weight is.

The notion of simple and unique structure is to identify the structures that may only be observed in a specific subset of the data. Unlike boosting, it is counterintuitive to use simple local structures to predict the entire dataset. To our knowledge, no previously proposed algorithm identifies the subsets with a simple structure for the data mining purpose. Usually, models try to fit close to data to achieve high accuracy, but there is also a need to understand how the model operates and makes decisions [18]. The boosting algorithms might achieve outstanding performance but suffer from a lack of model explainability [19].

Our algorithm aims to identify a limited number of subsets with simple structures; hence the simple model learned on each subset can explain the decision taken by the model. Unlike instance selection in our algorithm, a single subset represents the whole dataset partially, and the union of all the subsets will represent the entire dataset. Each subset will have a population of similar instances with unique and simple structures. The expected benefit is improved performance without using complex models such as support vector machines, random forest, neural networks, etc.

## I. Sub-setting Algorithm

In this section, we present the sub-setting algorithm. We offer the algorithm to identify subsets with simple structures and improved performance for the instances belonging to the subset. If the complexity of the data was well understood, one could construct suitable training data for the predictive model, but such understanding is absent most of the time; in such cases identifying similar instances is helpful. Identifying subsets with simple structures is complex; identifying such structures is by identifying subsets with similar instances. Similarly, the intuition behind similar instances is that some of the complex target–variable interactions are the same across the subset of similar instances; thus, the classification task is left to few less complicated relationships.

In the non-parametric $K^{th}$ nearest neighbor (KNN) algorithm, the classification of a query is achieved based on the class of the K nearest neighbors [20]. Because of the non-parametric nature of the KNN algorithm, it does not attempt to learn any underlying data distribution; instead, it assumes that similar instances are present in the close vicinity of each other. Thus, our sub-setting algorithm uses KNN space computed using Euclidian distance to identify the subsets of similar instances. Suppose we have training dataset $\boldsymbol{T}$ with $\boldsymbol{C}$ classes and m instances. $\boldsymbol{T}$ consists of $(\boldsymbol{x_i})_{\boldsymbol{i\epsilon[1,m]}}$ instances and a set of $\boldsymbol{f}$ features. To measure the similarity or distance between a query instance $\boldsymbol{q}$ and instance $\boldsymbol{x_i}$ Euclidian distance is one of the most widely used metrics given by (2.1) [21]

$$\boldsymbol{dist(q_f, x_{if})} = \sqrt{\frac{\sum_{j=1}^{f}(q_j - x_{ij})^2}{|f|}} \quad (2.1)$$

The similarity between two instances is inversely proportional to the Euclidian distance between two instances. Therefore, in KNN space, an instance's neighbors will be the most similar instances for it.

The remaining part of the section explains the details of the algorithm.

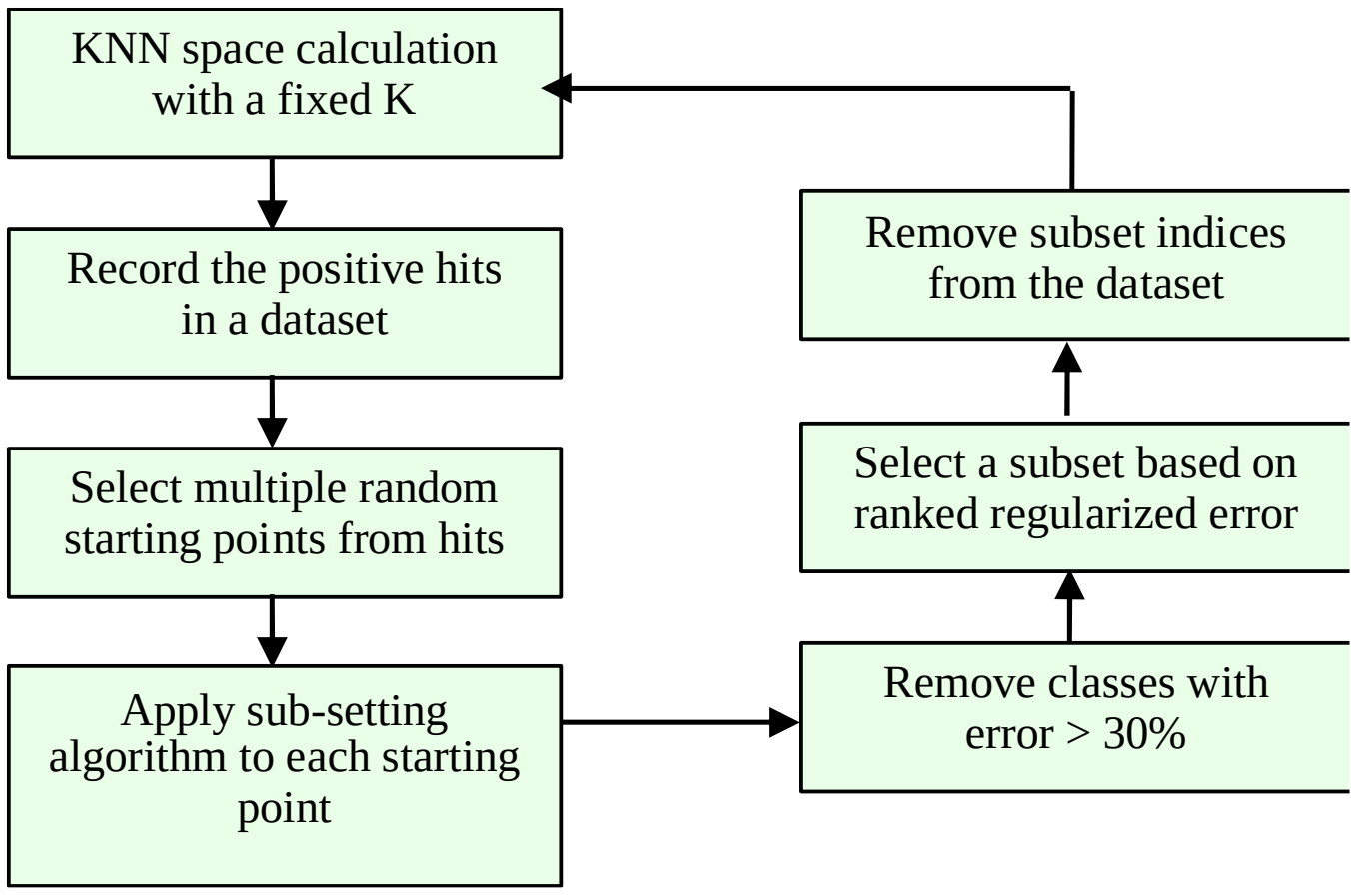

Fig. 2.1 – Sub-setting algorithm flowchart

Our sub-setting algorithm identifies $\boldsymbol{n}$ subsets from the training dataset $\boldsymbol{T}$; $\boldsymbol{S_i}$ represents a subset with $\boldsymbol{C'i}$ classes and $\boldsymbol{m'i}$ instances in each subset where $\boldsymbol{i}$ is the index of a subset. Remember for each subset Si,

$$\boldsymbol{m'_i < m} \ and \ \boldsymbol{C'_i \leq C} \quad (2.2)$$

$$S_1 \cup S_2 \cup S_3 \dots\dots \cup S_n = T \quad and \ (S_i)_{i\in[1,n]} \cap (S_j)_{j\in[1,n]} = \phi \qquad (2.3)$$

The (2.2) makes sure that the number of instances in a subset is less than the whole dataset instances, and the number of classes in a subset can be less than or equal to classes in the entire dataset. The (2.3) makes sure that the union of all the subsets forms a whole dataset, but there are no common instances between any two subsets.

Fig. 2.1 shows the overview of the algorithm. First, the algorithm computes the KNN space and fits a KNN model to the given training data. Our algorithm looks for the subsets incrementally, starting with a single instance. The algorithm predicts the training data, records the positive hits, and randomly selects an instance from the pool of positive hits as a subset seed. As similar instances are in the proximity of the subset, the seed is further grown by searching its K nearest neighbor, and we eliminate any misclassified neighbors (elimination step). The subset is further expanded by identifying $K^{th}$ nearest neighbor of new instances in the subset, followed by the elimination step. Every time the algorithm finds K new neighbors, some may be already present in the subset. The subset growth converges automatically when there are no new neighbors found. Once the subset is converged, the further iteration will give the instances already present in the subset. After identifying the first subset, the algorithm again fits the KNN model, identifies positive training hits for the new seed, and repeats the steps but excludes instances from found subsets from the training set. We stop subset identification when we have addressed the pre-determined proportion of the whole dataset using subsets.

The illustration of the algorithm is provided in fig. 2.2 We start with subset seed A and k = 3. We find its three neighbors B, C, and D. We eliminate B as the KNN model misclassifies it, then we find neighbors of new instances C and D and keep growing the subset. We stop the algorithm in the 4th step as the subset is converged, and no new members are found. So, the unique instances A, C, D, C2, C3, D1 form the first subset.

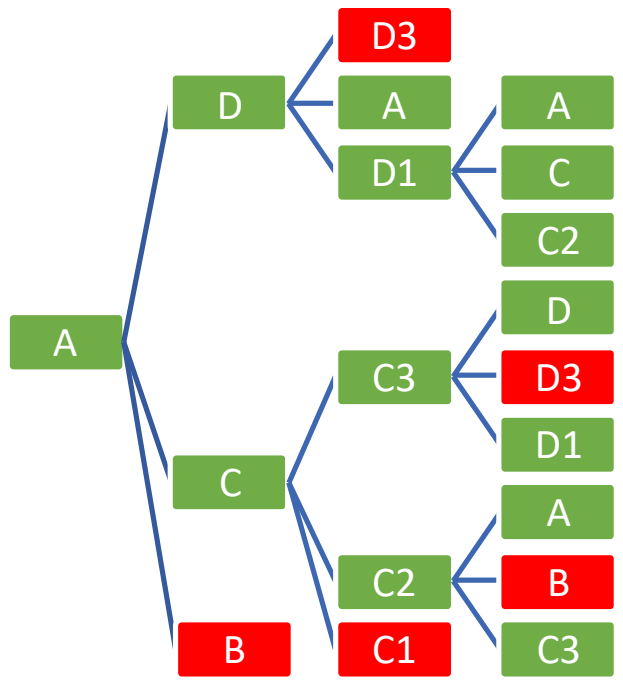

Fig. 2.2 – Algorithm illustration with k=3 (Instance A is seed. Green instances are positive hits and red instances are negative hits)

The growth of a subset converges automatically because of the existence of the natural clusters with similar instances in the Euclidian space. This is like identifying the clusters using the K-means clustering algorithm. K-means is an unsupervised clustering algorithm that divides the n data points into k disjoint clusters where identical data points are in the same cluster [22]. The quality of k-means clusters depends on the initialization points of the clusters; a poor initialization may result in a cluster stuck in the local minima, which is the case most of the time [23]. Similarly, the subsets formed by our sub-setting algorithm are sensitive to the starting point. The subset size, number of classes in the subset, and accuracy depend upon the subset initialization point. Some starting points may result in a tiny subset with a single class and very high accuracy; on the other hand, some initialization may result in a large subset consisting of more than 40% of data with diverse classes and low validation accuracy. The clustering task can be improved by selecting better starting points or initializing the clustering algorithm multiple times with different initialization points [22]. The vanilla K-means clustering algorithm is repeated several times, and clusters with minor loss are kept [24].

Similarly, the sub-setting algorithm doesn't want to miss any good subsets; it starts with multiple starting points and chooses the best subset. We use 10% of positive hits as potential seeds and identify subsets. As our subsets are supervised, we use the KNN training accuracy of the individual subset to choose the best subset from multiple initializations then repeat the steps to find subsequent subsets. Even though we have many starting points to identify a subset, a tiny subset with few instances always stands out because of very high training accuracy. So, we introduced regularization to penalize subset size. The regularization introduces smoothness to the learning function as smoothness is required to generalize the learned function [25]. We punish tiny subsets because of training impracticability and, on the other hand, penalize large subset sizes for poor validation performance. In regularization, small subsets lack generalization as the identified simple structure is common in very few instances; contrary, the large subsets are most general, indicating underfitting. Thus, we penalize both small and large subsets linearly according to (2.4)

$$Penalty = \begin{cases} \alpha l * (SST - m'i) & m'i \leq SST \\ \alpha u * (m'i - SST) & m'i > SST \end{cases} \quad (2.4)$$

In the (2.4) subset size threshold (SST) is a tunable user-defined desired subset size. The SST is dependent on the size of the training data. The subset size above and below SST will be penalized linearly and scaled down to match the algorithm error. If αl < αu shows that user desires smaller subset size and αl > αu shows inclination towards larger subsets. In fig. 2.3 as regularization has $SST$ 250, the penalty for subset size 250 is 0. As $\alpha l < \alpha u$ the $m'I > SST$ are penalized harshly than the $m'i \leq SST$. Because of this algorithm will have a bias to choose subset $m'i \leq SST$.

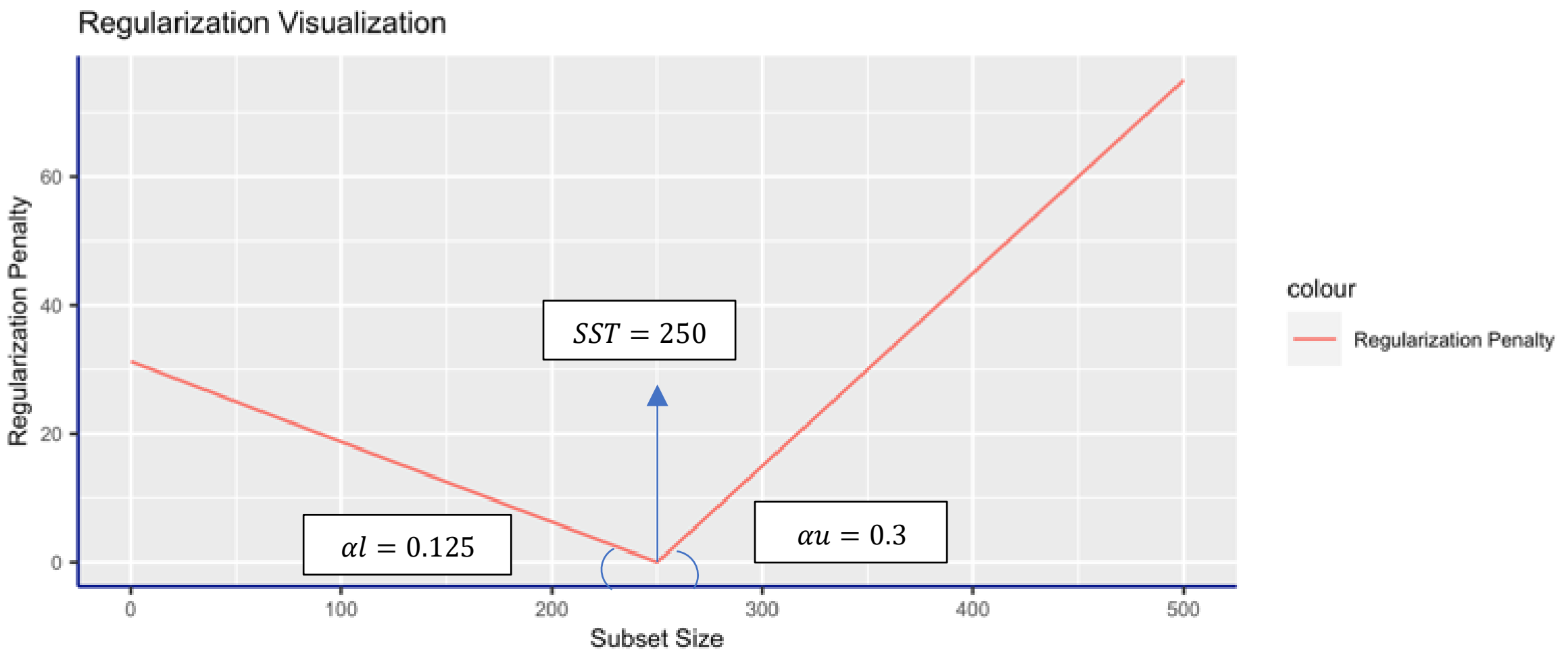

Fig. 2.3 – Regularization Visualization with SST = 250, αl=0.125, αu=0.3

The sub-setting algorithm starts with a predefined value of k but after addressing a particular portion of the data value of k should be increased. At the start of the algorithm, the data is very dense, so a smaller value of K will identify subsets with the given SST. As more subsets are identified in a further iteration, the data gradually becomes sparse, so existing K will result in a small subset. Hence, we increment the value of K to obtain the subset of the desired size with consistency.

## II. Data

We used the International Stroke Trial dataset (IST dataset) to test our algorithm [26]. This dataset is one of the most extensive randomized trials ever conducted in acute stroke. It contains three types of strokes Ischaemic Stroke, Haemorrhagic Stroke, Indeterminant Stroke. The goal was to predict the probability of survival after six months of stroke. We binned the available survival probability range into four equal-sized interval bins, with class 1 was the lowest probability class and 4 had the highest probability of survival. We ignored all the predictors collected after the discharge of the patient. We aggregated some features to create new features such as the medication, amount of medication, location for the medication injection, surgeries performed, etc. There are around 17300 data points in the dataset. There exist three patient categories in the dataset based on patient condition when admitted into hospital - Drowsy, Conscious, and Unconscious. We choose a drowsy state and apply our sub-setting algorithm to this type of patient. There are 3013 data points and 20 features.

## III. Experiments and Results

In this section, we provide the performance of our sub-setting algorithm on the "drowsy" condition of international stroke data and explain the need for the regularization and the effect of the different regularization parameters on the quality of the dataset. The algorithm is implemented in R using the Kernel KNN package [27] to compute KNN space and KNN implementation from caret (KNN3) [28] to fit the KNN model.

Table 4.1 shows the first 5 subsets identified by the algorithm without any regularization. The subsets in table 4.1 have perfect accuracy, but subset size is inadequate and impractical for training a data mining algorithm. The sub-setting algorithm uses accuracy and ignores the subset size as the only criteria for selecting the best subset in each iteration. In machine learning and data science, the algorithm overfits the data when it fits a model too close to data. As a result, the model performs well in training but worsens in the validation [29]. In our case algorithm overfits the data if it identifies the small subsets with perfect accuracy. The identified simple structure in the smaller subset is dominant across a small population; hence it will be rare to locate a new instance belonging to the pattern diminishing performance in the inference. Contrary, a large subset size will underfit the data as identified structures will be too general as it is counterintuitive to identifying different subsets with different simple and unique patterns. Hence to achieve a balance between large and small subsets, we use regularization as described in (2.4)

Table 4.1 – Top 5 subsets identified by the algorithm

| Subset Size | Number of Classes | Accuracy |
|---|---|---|
| 35 | 2 | 100 |
| 63 | 3 | 100 |
| 105 | 3 | 100 |
| 41 | 2 | 100 |
| 36 | 2 | 100 |

The KNN model cannot validate the subset quality as it has a learning bias towards the identified subsets. The KNN model will always perform well on the subsets as misclassified instances by the KNN model are removed by an algorithm to identify similar instances. To obtain unbiased subset quality, separate decision trees are learned on each subset, and these learned trees can be used to predict similar instances belonging to the respective subsets. The other benefit of the decision tree is nodes of the learned decision trees can reveal the different structures in the subsets. As there are no separate testing sets, 5-fold cross-validation is used to evaluate the actual accuracy of the subsets.

To avoid the poor subset size, the algorithm uses regularization described in (2.4) with $SST = 250$, $\alpha l = 0.125$, $\alpha u = 0.2$, and $k=3$ throughout the algorithm. Fig 4.1 shows the subset size of all the subsets in the order they are found by the algorithm. The first few identified subsets have desired subset sizes, but as more and more subsets are identified, the subset size reduces drastically regardless of the regularization penalizing smaller subset sizes. The subset sizes after the $10^{th}$ subset drop below 100, and the subset size after the $20^{th}$ subset are practically untrainable. After initial iterations, it seems that regularization is not helping the algorithm avoid a smaller subset size. According to table 4.2, in iteration 15 highest subset size identified by the algorithm is 50, and regularization is done, helping to avert tiny subsets. As overall determined subset size is poor, the algorithm with regularization attempts to select best from the poor.

Table 4.2 – Top 3 and bottom 3 subsets ranked by subset size in 15th iteration
(Accuracy is for the KNN model)

| Top 3 | | | |
|---|---|---|---|
| **Subset Size** | **Number of Classes** | **Accuracy (%)** | **Regularized Error** |
| 50 | 4 | 98 | 27 |
| 49 | 2 | 97 | 27 |
| 47 | 3 | 95 | 29 |
| **Bottom 3** | | | |
| 4 | 1 | 100 | 375 |
| 5 | 2 | 100 | 300 |
| 6 | 2 | 100 | 250 |

According to fig. 4.1 as the amount of data the subsets address increases, the subset size starts deteriorating and becomes untrainable after a certain point. The proportion of data addressed at $10^{th}$ subset 50 % and $30^{th}$ subset 65% as a result, remaining KNN space becomes sparse as more and more subsets are identified gradually by the algorithm. Hence, we need to adapt the value of K to the KNN space. The value of K in the algorithm represents the measure of similarity between instances. The similarity of the instances is inversely proportional to the value of K as K increases, the similarity between instances decreases. We start the algorithm with stringent similarity criteria ***K=3***, but after identifying the first few subsets, we are left with the instances rejected by the algorithm in the previous iterations. The instances rejected in previous iterations lack the similarity among themselves. Therefore, after initial iterations for stringent similarity constraints like the **K=3** algorithm starts identifying small size subsets with few very similar instances. If the algorithm increases the value of K, the Euclidian distance of the farthest neighbor increases, reducing the similarity among the instances. The increased value of K reduces the instance similarity requirement and focuses on maintaining the subset size by incorporating less similar instances from the neighborhood; thus, we increment the value of K gradually along with the algorithm iterations.

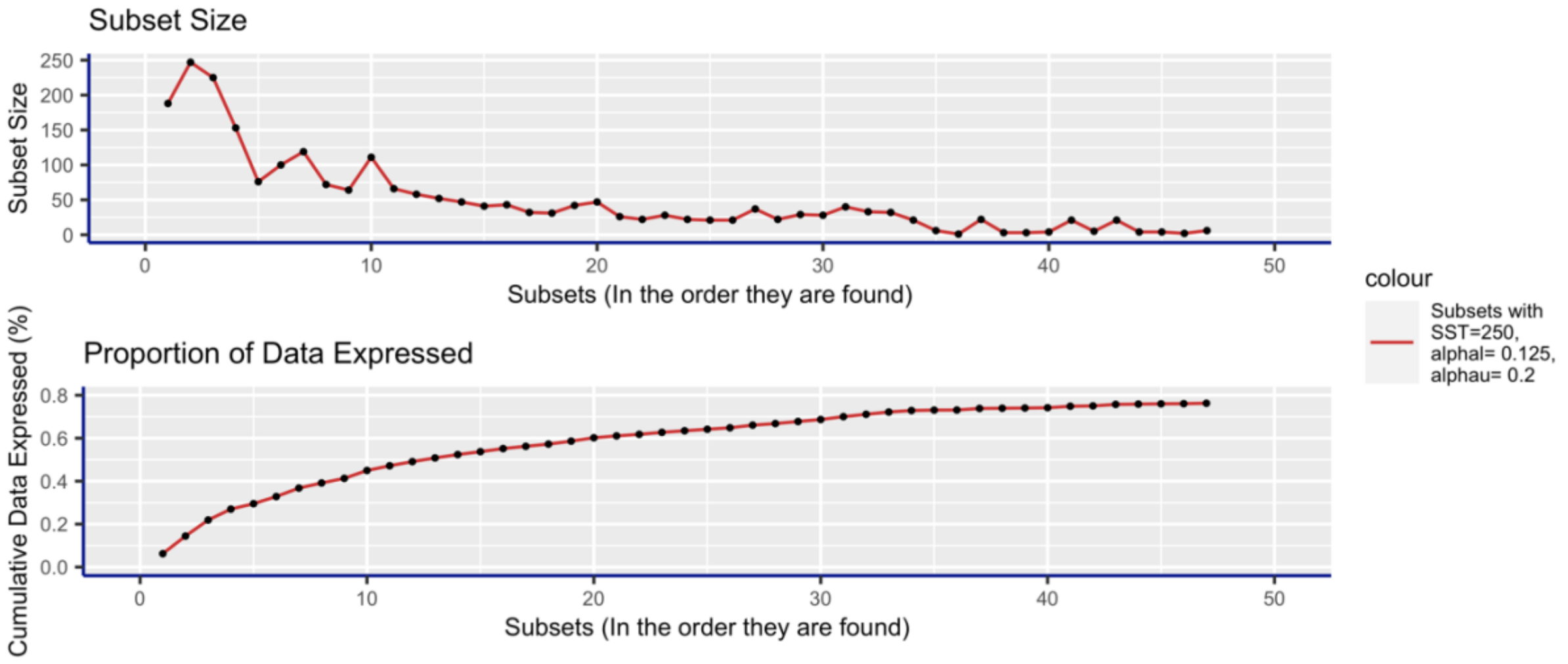

Fig. 4.1 Subset size for subsets with SST 250, $\alpha l = 0.125$, $\alpha u = 0.2, k = 3$

The fig. 4.2 shows 5-fold CV accuracy for subsets with ***SST 250,*** $\boldsymbol{\alpha l = 0.125, \alpha u = 0.3}$ The algorithm was started with ***K= 3***, and K was incremented to 5, 7, 9 after addressing 25%, 70%, and 85% of data, respectively. The CART decision trees with ***cp (Complexity Parameter) = 0.01*** were used to validate the subsets. The regularization penalizes small subset size heavily (see fig 2.3), but subset sizes close to SST are punished leniently. As a result, the subset size hovers between 100 and 300. The mean CV accuracy of the subsets is 64.89%, which is better than the whole dataset CV accuracy of 50%. Thus, along with regularization choice of K also affects the subset quality.

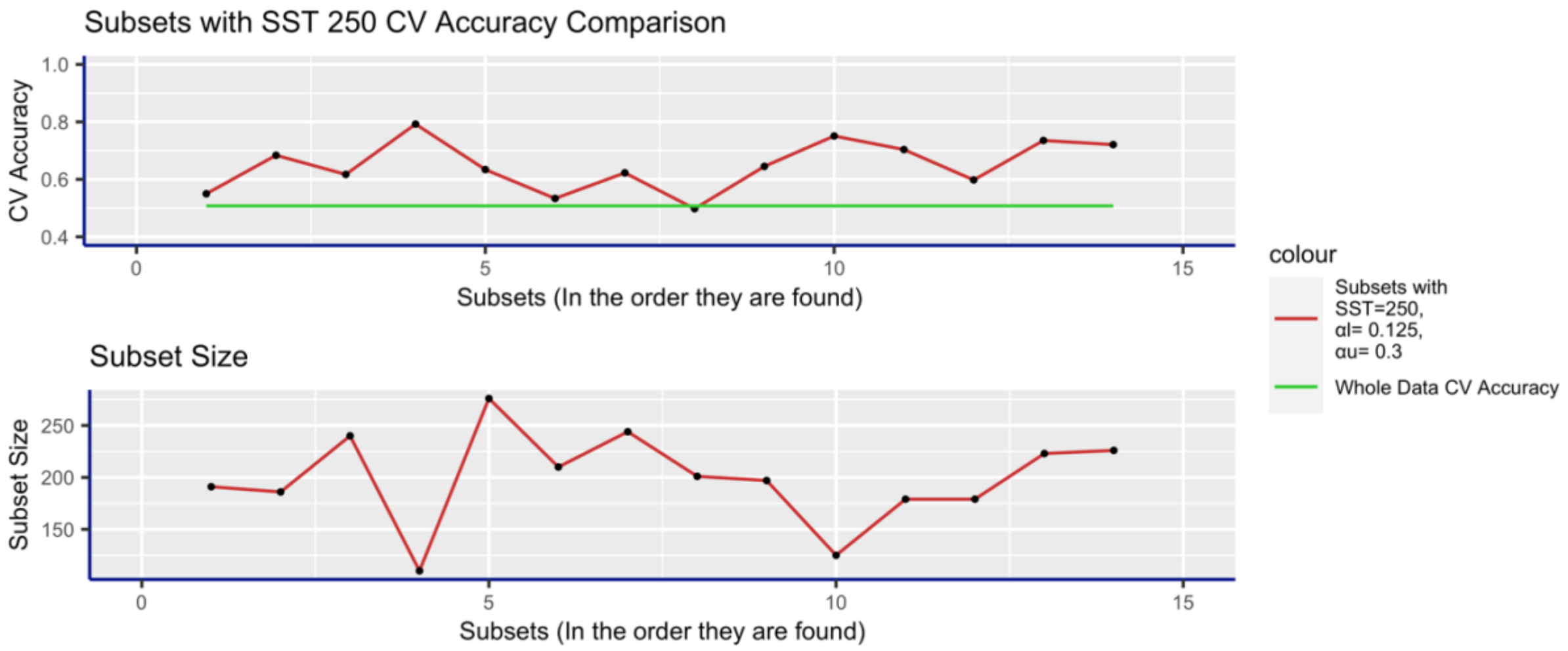

Fig. 4.2 Subset cross-validation accuracy (CV) comparison and subset size comparison for subsets with SST 250, $\alpha l = 0.125, \alpha u = 0.3$

The value of the subset size threshold (SST) determines the size of the identified subsets. The regularization prefers the subset size in the proximity of the SST value, whereas it avoids the subset size far from the SST by punishing them. Along with SST, proper choice of K is also vital to achieving the desired subset size. Fig. 4.3 shows the subsets with three different SSTs 150, 250 & 350. The values of K αl and αu are adapted to achieve a consistent subset size. For both ***SST = 150*** and ***SST = 250***, the algorithm started with K = 3, and K was increased to 5, 7 and, 9 after addressing 25%, 70%, and 85% of data, respectively. To achieve a larger subset size in ***SST = 350***, algorithms were started with K=5, and K was incremented to 7 and 9 after addressing 70% and 85% of data.

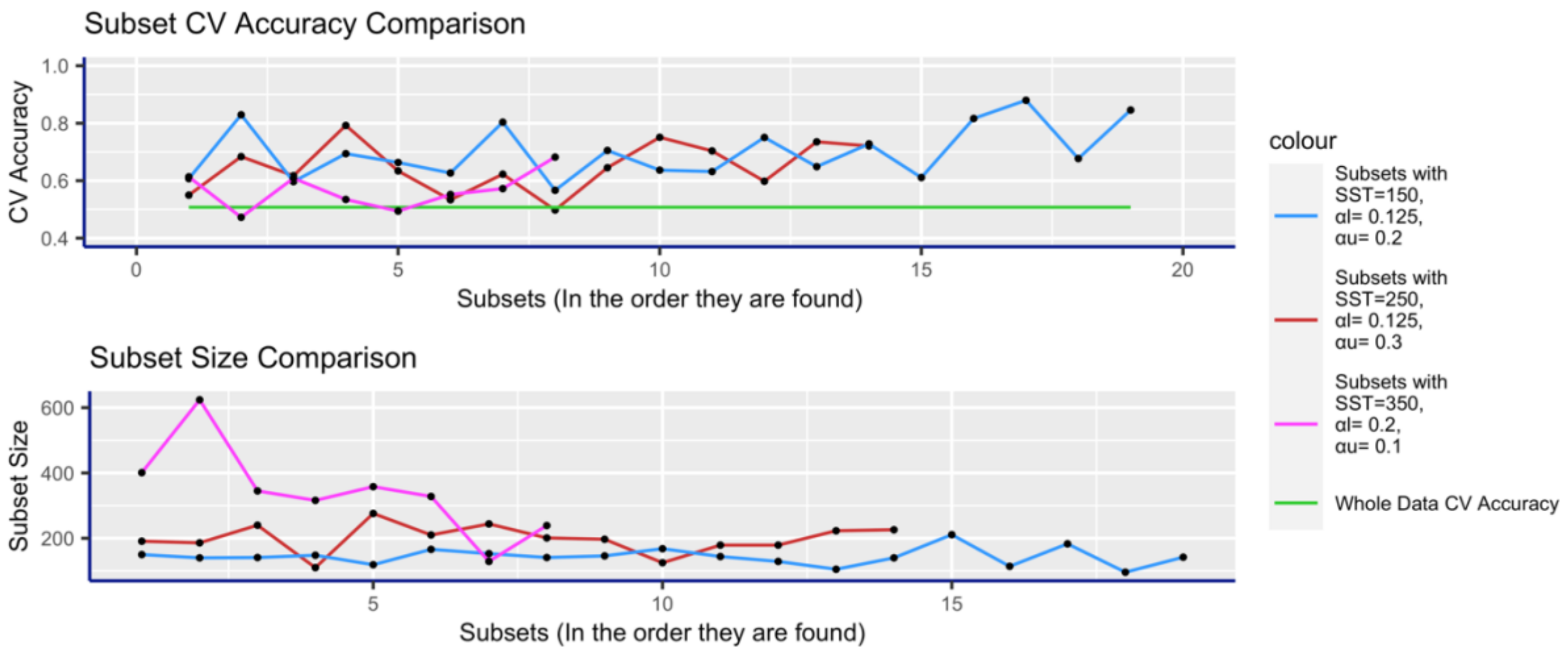

Fig. 4.3 Subset cross-validation accuracy (CV) comparison and subset size comparison for subsets with SST 150, SST 250, SST 350

Table 4.3 summarizes the performance of the different subsets with different SSTs. The subsets with SST 150 perform best with a mean CV accuracy of 70%, whereas those with SST 350 perform worse. From an accuracy point of view, subsets with SST 150 appear best, but the CV accuracy has the highest variance, and it requires 19 subsets to address 90% of data compared to 14 subsets for SST with 250. The algorithm is overfitting for SST = 150 as it identifies many small subsets with high mean accuracy. These small subsets might have poor generalization as the identified patterns are prominent in a small bunch of instances. As the subset size decreases, the number of identified subsets to address the same amount of data grows. The increased number of identified subsets poses a challenge of assigning a new datapoint to a suitable subset. Though this paper doesn't address the challenge of setting a new instance to one of the identified subsets, several subsets increase the probability of misclassifying a new instance into one of the subsets. The high variance associated with the subset (fig 4.3) accuracies for SST = 150 shows that the high mean accuracies of the subsets might be deceptive as some do exceptionally well while some don't. On the contrary, the mean accuracy of the subsets with SST = 350 is inferior (see table 4.3). The low mean accuracy indicates large subsets suffer from underfitting because the algorithm introduces less similar instances in the subsets to satisfy the subset size requirement, creating overly broad ones.

Table. 4.3 Summary statistics for subsets with SST 150, SST 250 and, SST 350

| Subset SST | Mean CV Accuracy (%) | Variance of CV Accuracy | Mean Subset Size | Number of Subsets to Address 90% of Data |
|---|---|---|---|---|
| **150** | 70.09 | 89.79 | 144 | 19 |
| **250** | 64.89 | 75.66 | 199.07 | 14 |
| **350** | 56.62 | 46.64 | 342.5 | 8 |

The small subsets are created by smaller values of the K and have more similarity in the instances resulting in high accuracy but training impracticability. The large subsets have a large value of K hence have less similarity among the instances resulting in poor performance. Therefore, we need to achieve a tradeoff between a large and small subset. Thus, subsets with SST 250 strike a good balance between particular and very broad subset patterns; it requires just 14 subsets to address 90% of data with a mean accuracy of 64%, whereas the accuracy of the decision tree on the whole dataset is just 50%. According to table 4.3, the subsets with the mean size of 200 address 6.5% of the entire data individually. To learn complex pattern data mining algorithms requires many data points, but if we want to learn a simple true linear relationship, just two instances are enough. Similarly, as we want to know the simple structure in a subset, 200 instances are enough to learn a structure. The performance of the subset is consistent as just two out of fourteen datasets are below 60%; other than that, all the subsets improve performance by at least 10%; additionally, the variance of the accuracy is lower than SST 150.

## IV. Discussion

The paper's premise is instances in a dataset belong to different data distributions; hence to make a successful prediction, we don't require all the datapoints; we just need datapoints belonging to the distribution of the query instance. Based on this assumption objective of this paper is to identify the subset of similar instances that might belong to the same distribution. Further, we argue that the identified subsets with similar instances have simple and previously unidentified structures; hence data mining algorithm will learn less complex mapping functions improving the performance of the data mining algorithm to predict a query belonging to the given subset. The sub-setting algorithm identifies a set of subsets that consistently perform better than the model trained on the whole dataset, provided a proper choice of SST and K to handle overfitting as well as underfitting.

The first obvious thing to verify is that subsets formed by the algorithm have a unique structure that helps subsets classify instances with better accuracy. The fig. 5.1 shows the comparison with accuracies of the random subsets and the subsets formed by the algorithm. Random subsets are not purely random. The size of the random subset is drawn from a normal distribution, and parameters of this distribution are derived from the distribution of the subset size found by the algorithm with SST 250. The CV accuracy of the subsets with SST = 250 is considerably above the CV of a random subset. The random subsets are even worse than the model trained on the whole dataset. This drastic difference in accuracy proves that the sub-setting algorithm takes advantage of the KNN space to identify similar instances and previously unidentified structures.

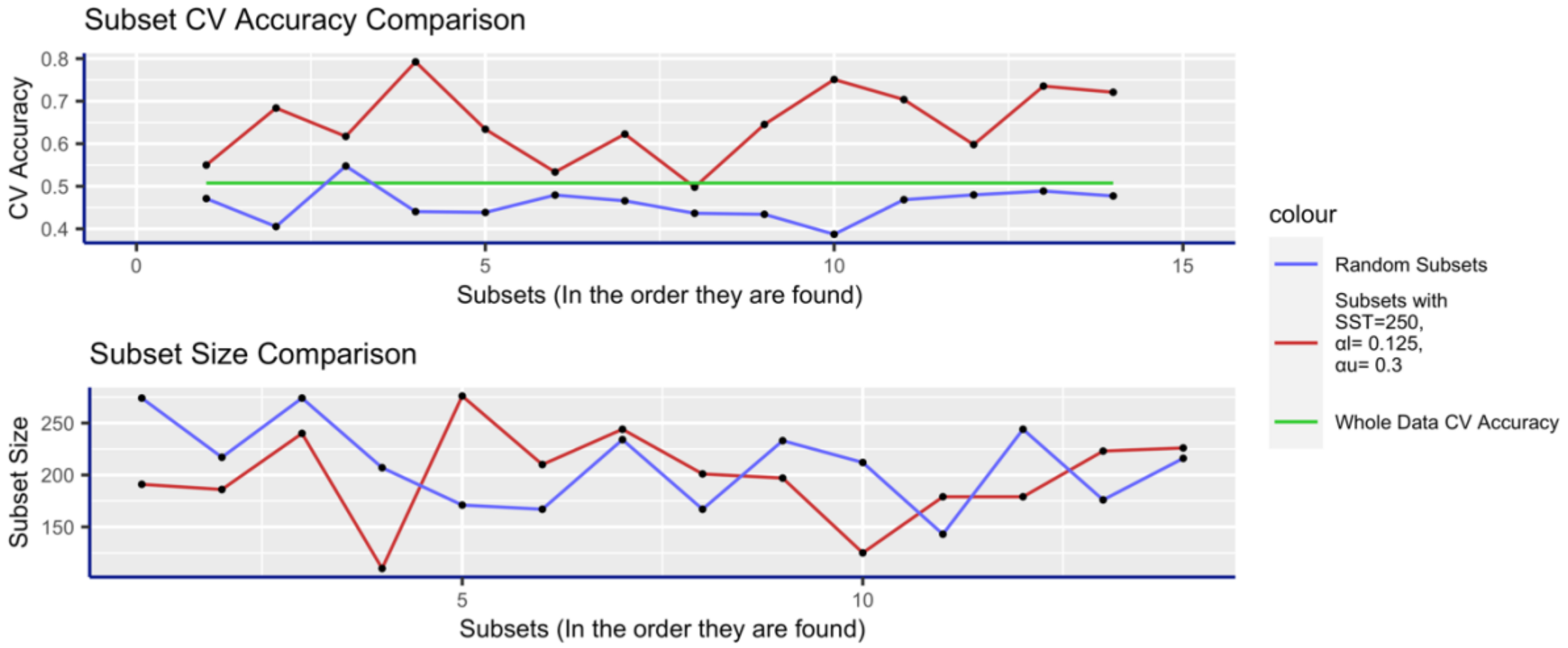

Fig. 5.1 Subset cross-validation accuracy (CV) comparison between random subsets and subsets with SST 250

In above section decision tree with default complexity parameter (cp) = 0.01 was used. The cp (complexity parameter) prunes the tree in the rprt library [30] of r. In a learned decision tree, if a split does not decrease the loss of the model by a factor cp, then that split is not attempted [31]. So higher value of cp prunes the tree by removing some of the nodes while lower values of cp grow bigger and complex trees which might overfit the data. Given the size of the subsets, the default cp value is very small, which leads to a complex tree structure with many nodes. Table 5.1 for cp = 0.01 whole dataset has 9 tree nodes and has CV accuracy of 50%; however, subsets identified by the algorithm achieve around 65% with a smaller number of nodes.

Table 5.1 – Performance of subsets (SST=250) with different complexity parameters (cp) of the decision tree

| CP | Whole Dataset DT Nodes | Whole Subset CV Accuracy (%) | Mean Random Subset DT Nodes | Subset Mean CV Accuracy (%) | Mean Subset DT Nodes | Random Subset CV Accuracy (%) |
|---|---|---|---|---|---|---|
| 0.01 | 9 | 50.74 | 11 | 64.9 | 7 | 45.8 |
| 0.015 | 6 | 48.48 | 9 | 65 | 7 | 46 |
| 0.02 | 3 | 47.5 | 8 | 65 | 5 | 46 |
| 0.025 | - | - | 5 | 64.9 | 4 | 46 |
| 0.05 | - | - | 2 | 64.4 | 2 | 44.9 |

Further in table 5.1, as cp is increased from 0.01 to 0.025, the number of nodes for the whole dataset reduces, so does the accuracy. This is expected as for higher values of cp, the decision tree learns less complex trees. But in the subsets identified by the algorithm, the subset accuracy stays the same regardless of the cp value. As the subsets identified by our algorithm have simple patterns, simple trees constructed using higher cp values do not degrade the performance; instead, they help overcome the overfitting caused by the complex tree over the small subset.

The subsets can be compared with whole dataset performance in 3 different ways 1) comparing trees constructed using same cp values 2) by comparing tress having same number nodes 3) best performing cp for both subset and whole dataset. As seen in Table 5.1, subsets identified by the algorithm do considerably better than the whole dataset model in all three cases. The subsets identified by our algorithm outperform the model learned on the whole dataset, especially when we force the model to learn a simple structure with a higher cp value. In table 5.1, the mean number of nodes and subset size of the random subsets and subsets identified by the algorithm are similar, particularly for higher cp values. Still, the subsets determined by the algorithm outperform the random subsets with a substantial margin. The performance of the random subsets is worse because the decision tree algorithm cannot construct meaningful decision trees on random subsets. On the other hand, the subsets identified by the sub-setting algorithm have unique structures that help them outperform the random subsets and models learned on the whole dataset.

Table. 5.2 – Features used as nodes in the decision trees learned on the subsets identified by the sub-setting algorithm

| Subset | RDEF1 | RDEF3 | RDEF4 | RDEF5 | RDEF6 | RDEF7 | RDEF8 | RATRIAL | RSBP | ASPI | HEP | Drug | STYPE |
|---|---|---|---|---|---|---|---|---|---|---|---|---|---|
| Whole | | | yellow | green | yellow | yellow | | yellow | | | | | |
| 1 | | | yellow | yellow | green | yellow | | | | yellow | yellow | | |
| 2 | | | | | green | | | | | | | | yellow |
| 3 | | | | yellow | | | yellow | yellow | yellow | | | | green |
| 4 | | | | | | | | | yellow | | | | green |
| 5 | | | yellow | green | yellow | | | yellow | yellow | | yellow | | |
| 6 | yellow | | yellow | yellow | green | yellow | yellow | yellow | yellow | yellow | | | |
| 7 | | | | yellow | | | | yellow | green | | | | yellow |
| 8 | | | | yellow | yellow | yellow | yellow | yellow | green | | | | |
| 9 | | yellow | | yellow | yellow | | | yellow | yellow | | yellow | | green |
| 10 | | | | | | green | yellow | yellow | | yellow | | | |
| 11 | | | yellow | yellow | yellow | | green | yellow | yellow | | yellow | | |
| 12 | yellow | | yellow | | green | yellow | | | yellow | | | | |
| 13 | | | yellow | green | | | | yellow | yellow | | | | |
| 14 | yellow | | | | yellow | yellow | | green | yellow | | | yellow | |

yellow: Node is Present    green: Root Node

In previous sections, subset accuracy and subset size were used to evaluate the quality of the subsets, but these methods don't assess the different structures present in these subsets. The goal of this algorithm is fulfilled only if subsets identify previously unidentified structures for the classification. The decision trees can be used to attest to the new structures existing in the subsets. Decision trees are in the form of nodes where each node split tries to minimize the entropy of the system [31]. The features which reduce the entropy of the data most are present in the learned decision trees. The structure of decision trees automatically gives a feature importance map where the node's root is the most critical node as it minimizes the entropy most and leaves are least vital as they reduce the entropy least. The algorithm has successfully identified previously unidentified structures if decision trees learned on subsets have different nodes than those discovered on the whole data set.

Fig 5.2 shows the decision tree learned on the whole dataset with cp = 0.015. Table 5.2 shows the different features of the decision trees learned on the subsets identified by the sub-setting algorithm. After comparing the nodes present in the subsets and the decision tree learned on the whole dataset, the subsets use some of the features unused by the whole dataset decision tree as nodes. The features STYPE (Stroke Type), RSBP (Systolic Blood Pressure), RDEF8 frequently appear in the structures identified by the subsets, but the whole dataset decision tree entirely ignores them. The top candidates for the primary split of the root node of the whole dataset decision tree are as follows –

RDEF5 splits as RLR, improve=116.73370, (0 missing)
RATRIAL splits as LR, improve=113.67240, (0 missing)
RDEF6 splits as RLR, improve=100.47950, (0 missing)
STYPE splits as LLLR, improve= 65.90203, (0 missing)
RDEF7 splits as RLL, improve= 62.68121, (0 missing)

The feature STYPE ranks 4th in the top candidates for the primary split for the root node of the whole dataset decision tree, but it never appears in the whole dataset decision tree. On the other hand, 6 out of 14 subsets use STYPE as one node, and 3 subsets use it as a root node. The RSBP is another vital feature identified by the algorithm, which appears twice as the root node and RDEF8 as once as a root node. Hence our algorithm has identified subsets with previously unidentified structures.

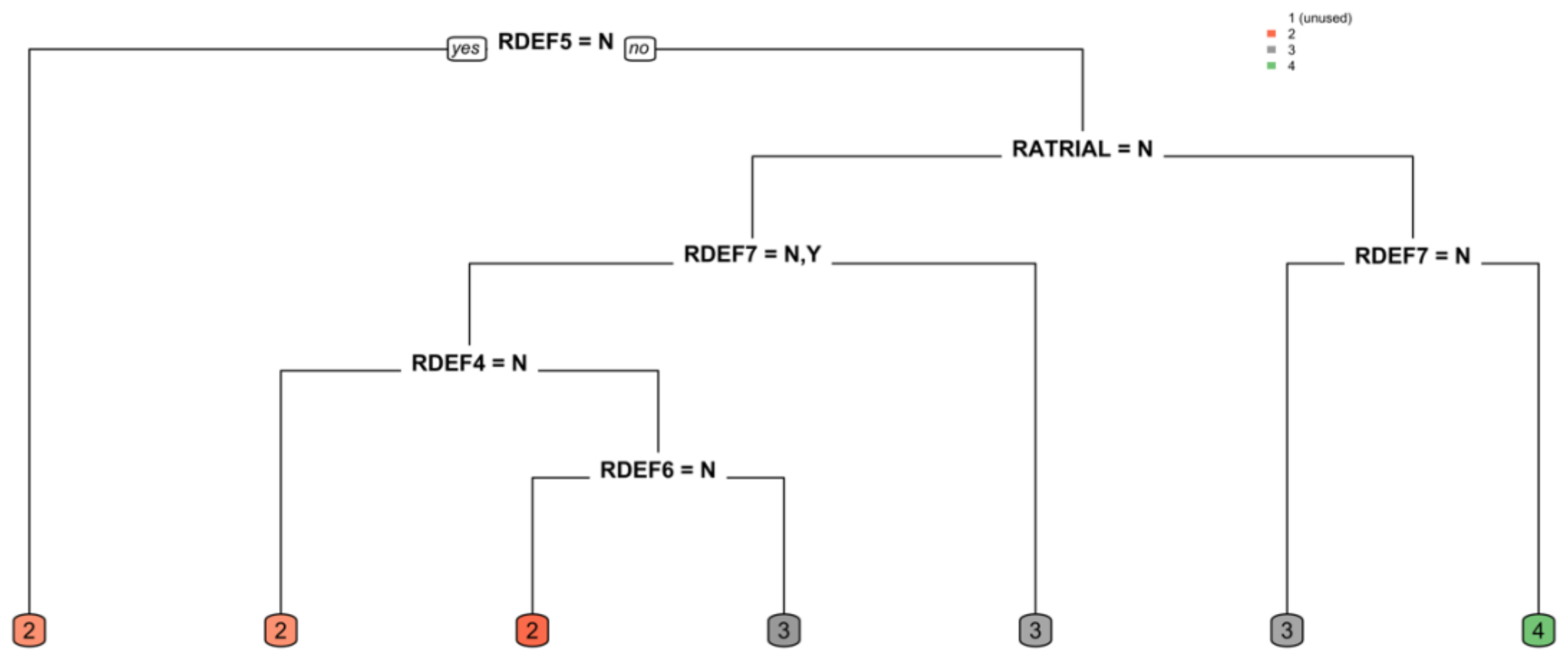

Fig. 5.2 Decision tree for the model learned on the whole dataset with cp = 0.015

The decision trees work in a top-down fashion. It divides the dataset into smaller subsets at each split. At the end of the decision tree algorithm, it divides the whole dataset into several subsets equal to the number of leaves. The decision trees split the data so that each split will reduce the entropy of the given node. On the contrary, our algorithm implements a bottom-up approach starting with a single instance and constructing a subset of similar instances. As a result of this bottom-up approach, the sub-setting algorithm can identify some of the valuable structures missed by the top-down decision tree.

Table 5.3 shows the performance of the subsets in terms of the accuracy of the decision tree learned on the subset, corresponding kappa, and 5 - fold cross-validation accuracy with cp = 0.015. The table also illustrates the performance of the whole dataset decision tree model to classify the instances in the subsets collectively. The accuracy and kappa value of the models learned on the subsets show that the unique pattern present in the subsets outperforms the general pattern leaned on the whole dataset.

The whole dataset model cannot take advantage of these structures as it can never collectively evaluate the instances present in the subset. As per the details provided in table 5.3, at the root node of the whole dataset decision tree, the instances get divided into 30% - 70% split almost always; hence whole decision tree model never evaluates these instances collectively. Kappa denotes the proportion of the classification which are not by random chance. In table 5.3, the kappa value of the subset classifier is consistently better than the whole dataset classifier. The subset gives us previously unidentified structures and consistent and reliable predictions according to the kappa value.

Further to identify whether the sub-setting algorithm has identified interesting structures, we can look at the composition of the subsets. Table 5.4 shows the number of instances per class and misclassified instances per class. The red row represents the number of misclassified instances if we use the whole dataset decision tree to predict the subset instances. The green row shows the number of misclassified subset instances by the model learned on the subset. The identified subsets can be classified into three groups. Group 1 represents the subsets that do consistently better than the model trained on the entire dataset. Group 2 represents the subsets that perform better than the whole dataset model overall but might be having deteriorated performance for some classes. In contrast, group 3 shows the subsets that perform equally to the entire dataset model or slightly worse.

Table. 5.3 – Subset performance

| Subset Number | Subset Size | No. of Nodes | Whole Dataset Decision Tree on Subset Instances | | Whole Dataset Model Accuracy on subsets | Whole Dataset Model Kappa on subsets | Subset Model Accuracy | Subset Model Kappa | 5-Fold CV Accuracy |
|---|---|---|---|---|---|---|---|---|---|
| | | | Left Split Proportion at Root Node | Right Split Proportion at Root Node | | | | | |
| Whole Data | 3013 | 6 | 0.28 | 0.72 | 0.51 | 0.25 | - | - | 0.48 |
| Subset 1 | 191 | 6 | 0.28 | 0.72 | 0.57 | 0.31 | 0.65 | 0.45 | 0.56 |
| Subset 2 | 186 | 2 | 0.24 | 0.75 | 0.66 | 0.4 | 0.7 | 0.41 | 0.70 |
| Subset 3 | 240 | 7 | 0.31 | 0.69 | 0.28 | 0.006 | 0.68 | 0.52 | 0.61 |
| Subset 4 | 110 | 3 | 0.31 | 0.69 | 0.61 | 0.39 | 0.79 | 0.61 | 0.79 |
| Subset 5 | 276 | 7 | 0.18 | 0.82 | 0.58 | 0.29 | 0.71 | 0.5 | 0.62 |
| Subset 6 | 210 | 11 | 0.25 | 0.75 | 0.62 | 0.34 | 0.7 | 0.51 | 0.52 |
| Subset 7 | 244 | 8 | 0.35 | 0.65 | 0.31 | 0.08 | 0.72 | 0.55 | 0.61 |
| Subset 8 | 201 | 9 | 0.35 | 0.65 | 0.41 | 0.17 | 0.62 | 0.47 | 0.48 |
| Subset 9 | 197 | 9 | 0.27 | 0.73 | 0.58 | 0.3 | 0.75 | 0.6 | 0.65 |
| Subset 10 | 125 | 4 | 0.10 | 0.90 | 0.36 | 0.03 | 0.84 | 0.65 | 0.75 |
| Subset 11 | 179 | 8 | 0.34 | 0.66 | 0.63 | 0.36 | 0.77 | 0.57 | 0.70 |
| Subset 12 | 179 | 6 | 0.30 | 0.70 | 0.65 | 0.33 | 0.76 | 0.53 | 0.62 |
| Subset 13 | 223 | 8 | 0.29 | 0.71 | 0.42 | 0.18 | 0.82 | 0.66 | 0.73 |
| Subset 14 | 226 | 7 | 0.26 | 0.73 | 0.58 | 0.37 | 0.79 | 0.64 | 0.71 |

Group 1 contains the subsets representing the minority classes and few subsets focused on specific classes of the dataset. In the drowsy condition of the stroke dataset, we have 4 classes, out of which class 1 and class 4 are minority classes, and the whole dataset model misses class 1 altogether and has very high errors for class 4. The sub-setting algorithm has identified many exciting subsets in which these minority classes have a minimal error rate, or these minority classes are majority classes. In Group 1 (see table 5.4), subsets 3 and 7 represent the subsets with the majority population from class 1, and subset 10 represents the subset with majority class 4. Thus, unlike the whole dataset decision tree, the decision tree learned on these subsets can classify classes 1 and 4 with low error. The fig. 5.3 shows the decision tree constructed for subset 7. The tree structure uses an entirely different set of features than the whole dataset decision tree (see fig 5.2). The pattern in subset 7 focuses on the RSBP (systolic blood pressure) and STYPE (stroke type) to classify the instances; thus, this pattern can predict class 1 instances with high accuracy.

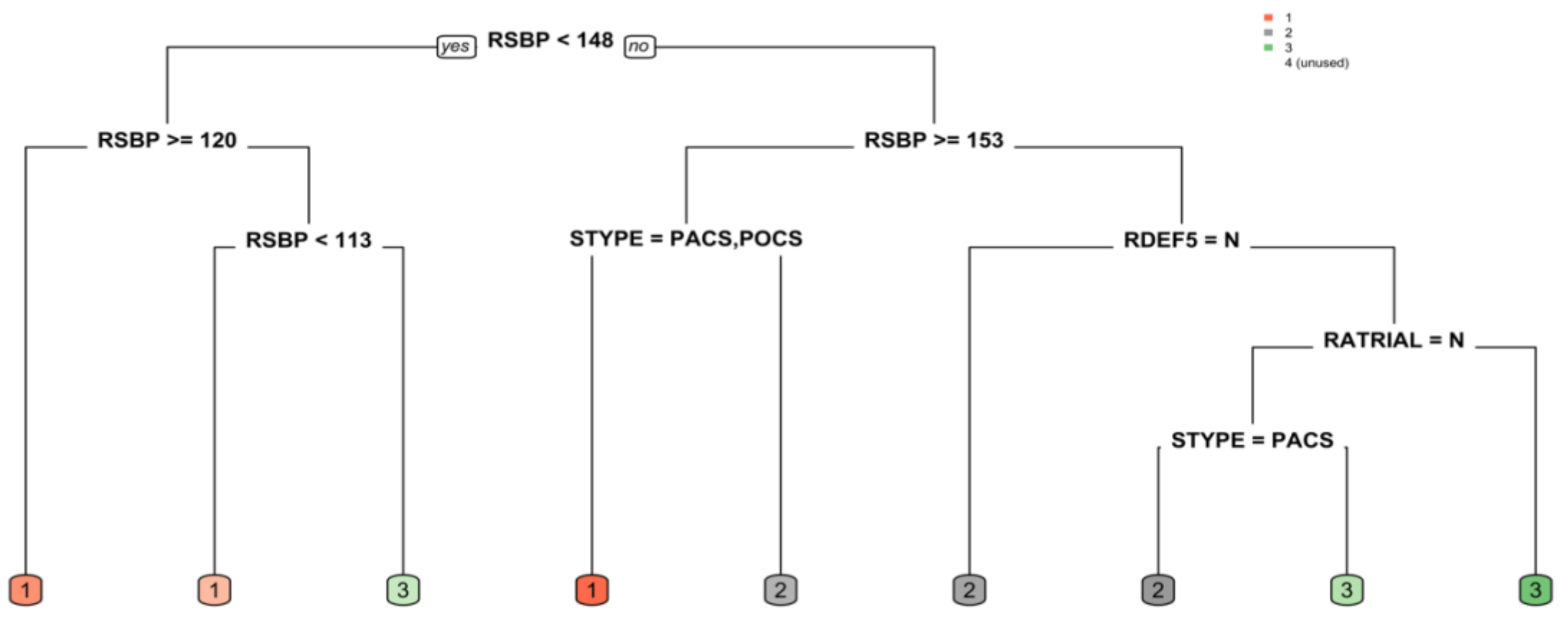

*Fig. 5.3 Decision tree for subset 7 with cp = 0.015*

In group 1, apart from subset 3, all the subsets are composed of at most three different classes instead of all the four classes. As subsets focus on specific classes or instances of the dataset, the classification model can learn simpler patterns with better accuracy. Subset 10 focuses on class 3 and class 4, whereas subset 12 primarily focuses on class 2 and 3. These subsets are concentrated on two classes in the dataset, and they yield simple structures with 3 and 6 nodes. Fig 5.4 shows the decision tree flow for subset 10. The identified subsets enable the predictive models to identify previously unidentified structures with different feature importance that drastically improve the predictive models' performance. RDEF7, the leaf node for the entire dataset model, serves as the root node for the subset 10 decision tree and uses new features aspirin and RDEF8, which were previously ignored. This previously unidentified structure improves kappa value from 3% (for whole data DT) to 65% (subset DT) (see table 5.3). The RDEF7 has the most differentiative power for class 4 instances; hence, the root node correctly classifies approximately 65% of the class 4 instances. The fundamental cause for the improved performance of the subset is reversed feature importance between RDEF7 and RATRIAL compared to the whole dataset decision tree (see fig. 5.4(a) and 5.4(b)). For the whole dataset decision tree, even though the root node splits most of the subset 10 on the right side of the split, enabling the decision tree to learn a pattern to classify the subset 10 instances, but the decision tree fails to do so because the top-down decision tree learns the splits which will minimize the entropy of the dataset globally. It looks like RATRIAL minimizes the entropy globally; hence it precedes the RDEF7, but as we evaluate subset 10 instances collectively, the RDEF7 reduces the entropy of the subset 10 most; hence for subset 10, RDEF7 precedes the RATRIAL. Therefore, the entire dataset decision cannot learn more specific but simple structures present in the dataset as they are local to a given set of instances.

In group two, according to table 5.4 for subsets 1 and 4, the overall performance of the subsets is improved compared to the whole dataset model, but the performance of the subset decision tree might deteriorate for the specific

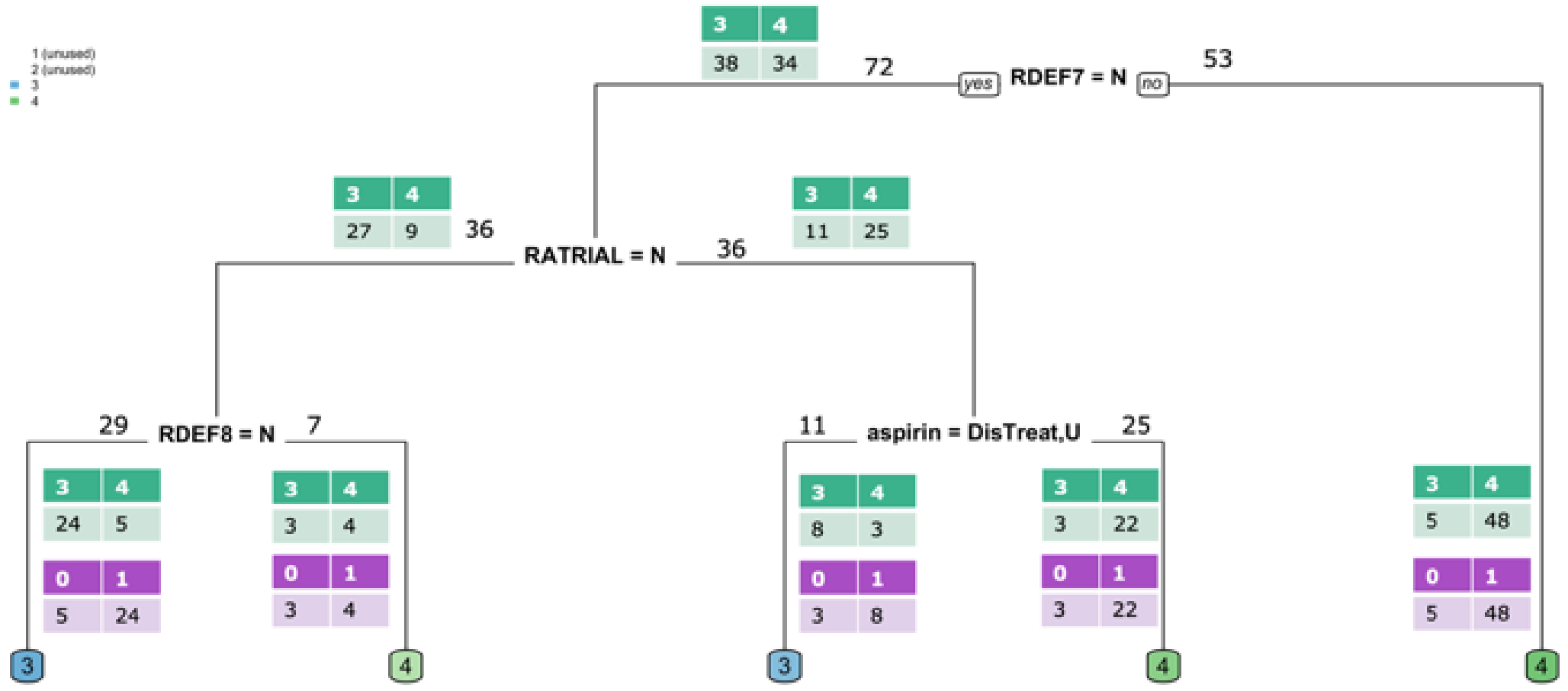

Fig 5.4 (a) – Subset 10 decision tree flow. The numbers at the left and right sides of the split show the split counts. The green table in the fig. indicates the number of instances per class at a given node, whereas the pink table indicates the number correct (1) and wrong (0) predictions

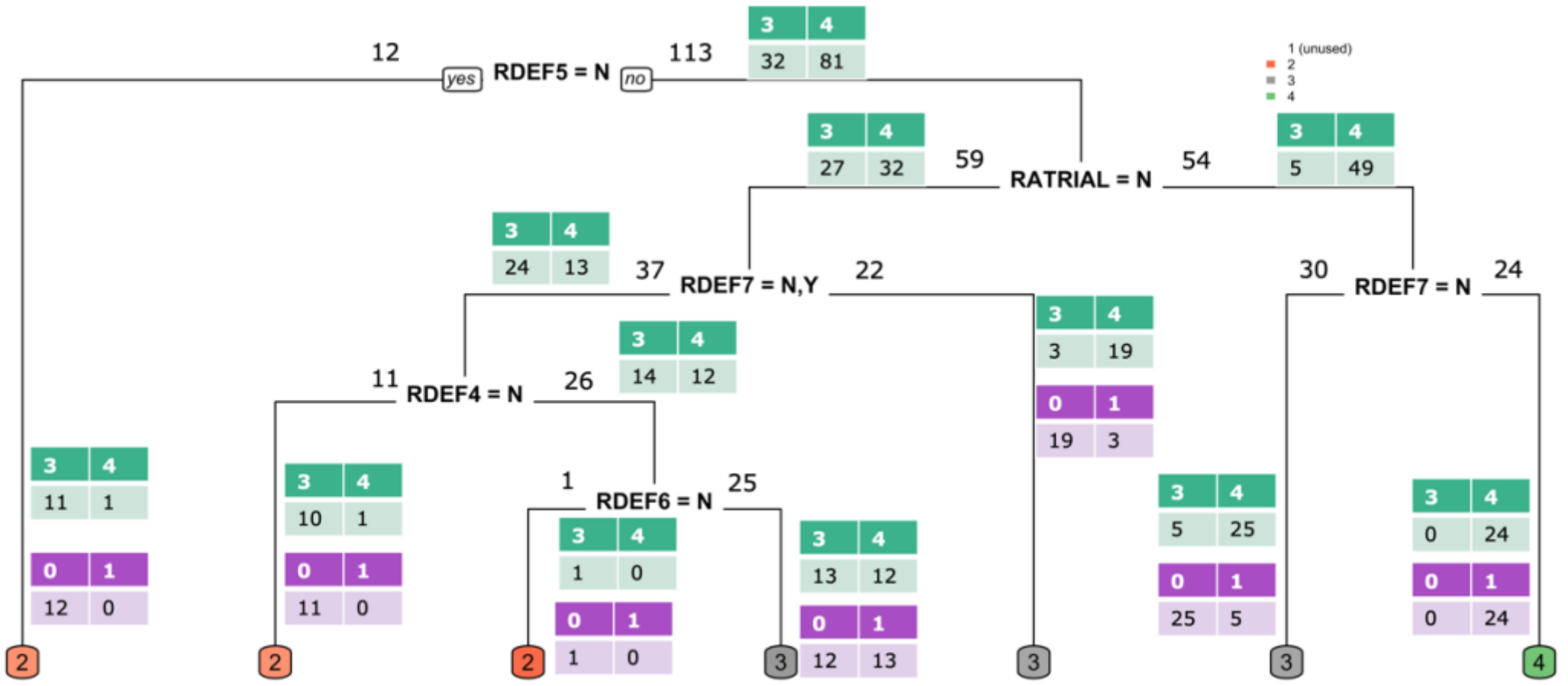

Fig 5.4 (b) – Subset 10 whole dataset decision tree flow. The numbers at the left and right sides of the split show the split counts. The green table in the fig. indicates the number of instances per class at a given node, whereas the pink table indicates the number correct (1) and wrong (0) predictions group of the instances. In the case of subset 4, the overall performance improved by more than 15%, accompanied by a substantial improvement in kappa, but if we look at the class accuracy, class 4 accuracy dropped to 0%.

According to table 5.3, the CV accuracy of subset 6 and subset 8 is the same as the CV accuracy of the whole dataset. Subset 6 has class 2, class 3 as majority class, and class 4 as minority class. This subset 6 pattern is like the pattern learned by the whole dataset decision tree; hence the performance of this subset is like a whole dataset decision tree. In subset 8, the model simultaneously addresses minority classes 1 and 4 and fails to do so, and it seems that the model can handle one of the minority classes at a time. Both the subsets 6 and 8 have a highly complex pattern with 11 and 9 nodes, respectively, which is contradictory to learning simple structures; hence they have poor validation performance. In short for few subsets such as subset 6 and subset 8 the model might try to capture specific unique pattern which looks promising from training error point of view, but such patterns might fail miserably in reality.

The learned subsets from the sub-setting algorithm address 90% of the data, and for the remaining 10 % of the data, the subsets formed are of tiny size and practically untrainable for a machine learning algorithm. The reaming 10% of the data cannot be treated as a simple structure subset because the algorithm does not identify it; hence logically, these instances will be best addressed by the general whole dataset decision tree.

The sub-setting algorithm improves the accuracy of the prediction model by identifying subsets with simple and previously unidentified structures; the model trained on such subsets classifies the instance belonging to the same subset with better accuracy than the general whole dataset decision tree. The improved accuracy for the instances is achieved at the expense of the computational cost and the complexity of the task. A top-down decision tree is the most efficient algorithm as it greedily splits the data to reduce the entropy of the data most. Even though the decision tree moves through the dataset multiple times, it evaluates the dataset collectively. But on the other hand, a bottom-up sub-setting algorithm looks for similar instances in the neighborhood of a given instance, and it evaluates a single instance at a time and moves through the data multiple times. The boosting algorithm might achieve better accuracy than our sub-setting algorithm but boosting lacks explainability. Compared to boosting, the sub-setting algorithm provides few subsets with a simple structure; as a result, it is possible to explain each subset of the algorithm. Thus, our sub-setting algorithm achieves a good tradeoff between explainability and accuracy. Our algorithm will fail if the dataset doesn't have local or straightforward structures, and there is no way to identify such a situation beforehand.

## V. CONCUSION

For classifying an instance, we don't require all the instances present in the dataset. An ensemble of the local patterns represents the data better than a single global mapping function. Given the proper choice of parameters, the sub-setting algorithm can identify subsets with local, simple, and previously unidentified structures. The identified subsets almost always perform better than the traditional pattern recognition algorithm; on the other hand, they also have better explainability than the complex pattern recognition models. Each subset identified by the sub-setting algorithm characterizes a distinct populous of the data. The bottom-up sub-setting algorithm incurs a high computational cost, unlike traditional top-down decision trees.

Table. 5.4 – Subset per class instance distribution

| Group 1 | | | | |
|---|---|---|---|---|
| Class | 1 | 2 | 3 | 4 |
| Subset 3 | 96 | 85 | 48 | 11 |
| Whole data DT misclass. | 96 | 52 | 16 | 8 |
| Subset DT missclass. | 11 | 39 | 23 | 2 |
| | | | | |
| Subset 5 | 0 | 63 | 150 | 63 |
| Whole data DT misclass. | 0 | 14 | 49 | 51 |
| Subset DT missclass. | 0 | 34 | 28 | 17 |
| | | | | |
| Subset 7 | 119 | 72 | 53 | 0 |
| Whole data DT misclass. | 119 | 35 | 12 | 0 |
| Subset DT missclass. | 16 | 41 | 11 | 0 |
| | | | | |
| Subset 9 | 33 | 78 | 86 | 0 |
| Whole data DT misclass. | 33 | 28 | 20 | 0 |
| Subset DT missclass. | 15 | 22 | 11 | 0 |
| | | | | |
| Subset 10 | 0 | 0 | 43 | 82 |
| Whole data DT misclass. | 0 | 0 | 22 | 56 |
| Subset DT missclass. | 0 | 0 | 11 | 8 |
| | | | | |
| Subset 11 | 0 | 104 | 49 | 26 |
| Whole data DT misclass. | 0 | 26 | 21 | 18 |
| Subset DT missclass. | 0 | 9 | 18 | 14 |
| | | | | |
| Subset 12 | 0 | 89 | 90 | 0 |
| Whole data DT misclass. | 0 | 28 | 33 | 0 |
| Subset DT missclass. | 0 | 20 | 22 | 0 |
| | | | | |
| Subset 13 | 69 | 0 | 127 | 27 |
| Whole data DT misclass. | 69 | 0 | 41 | 19 |
| Subset DT missclass. | 22 | 0 | 6 | 11 |
| | | | | |
| Subset 14 | 0 | 123 | 36 | 67 |
| Whole data DT misclass. | 0 | 41 | 14 | 38 |
| Subset DT missclass. | 0 | 14 | 24 | 8 |

| Group 2 | | | | |
|---|---|---|---|---|
| Class | 1 | 2 | 3 | 4 |
| Subset 1 | 11 | 74 | 79 | 27 |
| Whole data DT misclass. | 11 | 25 | 26 | 19 |
| Subset DT missclass. | 11 | 17 | 26 | 12 |
| | | | | |
| Subset 2 | 9 | 61 | 103 | 13 |
| Whole data DT misclass. | 9 | 16 | 28 | 9 |
| Subset DT missclass. | 9 | 29 | 3 | 13 |
| | | | | |
| Subset 4 | 18 | 49 | 37 | 6 |
| Whole data DT misclass. | 18 | 17 | 7 | 0 |
| Subset DT missclass. | 8 | 8 | 1 | 6 |

| Group 3 | | | | |
|---|---|---|---|---|
| Class | 1 | 2 | 3 | 4 |
| Subset 6 | 0 | 88 | 95 | 27 |
| Whole data DT misclass. | 0 | 28 | 30 | 21 |
| Subset DT missclass. | 0 | 26 | 24 | 11 |
| | | | | |
| Subset 8 | 43 | 69 | 51 | 38 |
| Whole data DT misclass. | 43 | 21 | 23 | 30 |
| Subset DT missclass. | 22 | 10 | 28 | 16 |